%% file: iclr2021_conference.tex
\documentclass{article} 
\usepackage{iclr2021_conference, times}

\input{math_commands.tex}

\usepackage{graphicx}
\usepackage{hyperref}
\usepackage{multirow}
\usepackage{url}
\usepackage{booktabs}
\usepackage{wrapfig}

\title{Technical report: Improving the properties of molecules generated by LIMO }

\author{Vineet Thumuluri\\
Computer Science and Engineering\\
  University of California San Diego \\
\texttt{vthumuluri@ucsd.edu}\\
\And
Peter Eckmann \\
Computer Science and Engineering\\
  University of California San Diego \\
  \texttt{peckmann@ucsd.edu}\\
\And
  Michael K. Gilson  \\
  Skaggs School of Pharmacy and Pharmaceutical Sciences\\
  University of California San Diego \\
  \texttt{mgilson@health.ucsd.edu} \\
\And
  Rose Yu  \\
  Computer Science and Engineering\\
  University of California San Diego \\
  \texttt{roseyu@ucsed.edu} \\
}

\iclrfinalcopy 
\begin{document}

\maketitle

\begin{abstract}
 This technical report investigates variants of the Latent Inceptionism on Molecules (LIMO) framework to improve the properties of generated molecules. We conduct ablative studies of molecular representation, decoder model, and surrogate model training scheme. The experiments suggest that an autogressive Transformer decoder with GroupSELFIES achieves the best average properties for the random generation task.
\end{abstract}

\section{LIMO framework}
LIMO (\cite{eckmann2022limo}) is a molecular generation technique that improves a given set of properties by mapping molecules to a latent space and uses inexpensive property surrogates to turn a discrete space optimization problem into a continuous one. There are multiple components to this framework which are described below.

\subsection{Generative model}
Given a dataset of molecular strings, a generative model is first trained to mimic the distribution of SELFIES \cite{Krenn2020} tokens, i.e. smallest units of a molecule string. In this case, a Variational AutoEncoder (VAE) \cite{kingma2022autoencoding}, is trained to map molecules ($X$) to a latent space ($Z$) using an encoder ($E$) i.e. $E: X\xrightarrow{}Z$ and decoder (D) that learns an inverse mapping i.e. $D: Z\xrightarrow{}X$, such that it is easy to reconstruct $X$ from Z using the decoder ($D$) i.e. $D(E(X))$ $\approx$ $X$ (Reconstruction), and the latent space is close to a unit normal distribution i.e. $D_{KL}(z || \mathcal{N}(0,\,I)) \approx 0$ (Regularization).

This defines a generative model $p(x) = \int p(x|z)p(z)$.  
 The VAE parameters are obtained by minimizing the ELBO loss \cite{kingma2022autoencoding}. In this work, we use a variant that weights the regularization term to avoid a problem during optimization where the KL term vanishes \cite{higgins2017betavae}.

$$
L_{\mathrm{BETA}}(\phi, \beta)=-\mathbb{E}_{\mathbf{z} \sim q_\phi(\mathbf{z} \mid \mathbf{x})} \log p_\theta(\mathbf{x} \mid \mathbf{z})+\beta D_{\mathrm{KL}}\left(q_\phi(\mathbf{z} \mid \mathbf{x}) \| p_\theta(\mathbf{z})\right)
$$

$\beta$ value of 0.1 was used in LIMO (\cite{eckmann2022limo}).

\subsection{Surrogate Model}
LIMO uses a property predictor model to map the molecular representation to property values \cite{eckmann2022limo}. The 
 molecular representation can either be a sample from the latent space or the decoder output. The property predictor acts as a fully differentiable and faster surrogate model to the oracle property predictor tools that are often computationally expensive. The properties are generally scalar values and thus the parameters of the neural network are learned by minimizing the mean-squared error between the predictions and oracle property values.

\subsection{Reverse optimization}
The key idea of LIMO is the latent inceptionism technique to reverse optimize the latent space and generate molecules with desired properties. Specifically, a large number of random points are sampled according to the unit normal distribution in the latent space. These serve as the starting points for the search. Molecules with the optimized properties are generated by local optimization in the latent space with the gradient approximated using the trained surrogate model. The overall objective being optimized is a weighted sum of one or more property values. All of the minima are then decoded to molecules and the top molecules, measured using the oracle property value predictor, are chosen.

\section{Attempted modifications to LIMO}
We conduct an extensive ablation study of  the LIMO framework, investigating different variations of LIMO components. These include changes to the tokenization of the molecular strings, the decoding model from the latent space to the molecule, and changes to how the property predictor is trained.

\subsection{Molecular Representation}
\label{molrep}
To represent molecules, LIMO used SELFIES \cite{Krenn2020}, a molecular string representation that ensures the tokens can be combined in any way to always form valid molecules.
The molecular strings can be decomposed into tokens in many ways.  GroupSELFIES  \cite{groupselfies2022} builds on top of the chemical validity guarantees of SELFIES by enabling group tokens, thereby creating additional flexibility to the representation. It extends  SELFIES where the tokens can be larger fragments and depend on the dataset and the fragmentation technique. 

In this work, we use the default fragmentation method from \cite{groupselfies2022} with different datasets:

(1) GS-Paper: GroupSELFIES tokens from the original paper \cite{groupselfies2022}.

(2) GS-USPTO: GroupSELFIES tokens extracted from USPTO reactions \cite{Lowe2017}.

(3) GS-Zinc: GroupSELFIES tokens extracted from ZINC250K molecules \cite{Akhmetshin2021}.

Table~\ref{tab:tokens} shows the total number of tokens, the average and maximum sequence length for each of the molecular string representations for the training dataset (ZINC 250K) considered in this study.

\begin{table}[h]
\centering
\caption{Statistics of different molecular string representation schemes when tokenizing the training dataset (ZINC250K).}
\label{tab:tokens}
\begin{tabular}{ccccc}
\toprule
\multicolumn{1}{l}{} & \multicolumn{4}{c}{Tokenizer} \\
\midrule
 & SELFIES & \multicolumn{3}{c}{GroupSELFIES} \\
\multicolumn{1}{l}{} &  & ZINC250K & USPTO & Paper \\
\midrule
Total tokens & 108 & 304 & 242 & 248 \\
Max length & 72 & 75 & 77 & 93 \\
Avg. length & 37.43 & 37.07 & 36.61 & 29.86 \\
\bottomrule
\end{tabular}%
\end{table}

\subsection{Decoder model}
The decoder in LIMO maps from latent space to a molecular string representation. Such mapping can be approximated in multiple ways. Here we discuss two broad strategies i.e. Non-autoregressive (NAR) and Autoregressive (AR) decoder models. In this work, both of these are implemented using the transformer architecture \cite{vaswani2023attention}.

\subsubsection{Non-autoregressive models}
LIMO \cite{eckmann2022limo} uses an MLP decoder by modelling the joint distribution of the target tokens ($y$) as conditionally independent i.e. $P(y) = \prod_{i=1}^n P\left(y_i \mid Z\right)$. While this has the advantage of inference being parallelizable, it suffers from the multi-modality problem \cite{gu2018nonautoregressive}. There have been multiple works that try to mitigate this issue through iterative refinement such as CMLM \cite{yang2021universal}, which uses masked language modeling and multi-step decoding. At inference time, first, the input is fully masked and in each subsequent iteration, the least likely tokens are masked and the procedure is repeated. A follow-up work CMLMC \cite{huang2022improving}, aims to improve the training objective of CMLM by addressing the mismatch at inference by including a token denoising loss in addition to the masking loss.
Another advantage of NAR models is the ability to do conditional generation by constraining on a given scaffold.

\subsubsection{Autoregressive models}
Autoregressive models, such as Transformers, are slower since only one token is decoded at each step and the cost scales at least linearly with the length of the target sequence length. However, the full joint distribution can be modeled i.e. $P(y) = \prod_{i=1}^n P\left(y_i \mid y_{1}, ..., y_{i-1}, Z\right)$, and thus does not suffer from the multi-modality problem. AR models thus have been shown to have superior task performance \cite{ren2020study}. In contrast to NAR models, conditional generation is non-trivial, thus losing the ability to do scaffold-constrained generation.

\subsection{Surrogate Model training}
The property predictor surrogate model training depends on the  VAE latent space i.e. the input to the surrogate predictor used in LIMO is a decoded representation of the point in the latent space, there are two approaches to optimizing the whole system. 

The first is to perform the training sequentially, in which case, the VAE is trained to convergence using the molecular string dataset and then its parameters are frozen. This ensures a fixed latent space. Random molecules are generated and their property values are computed to create the training dataset for the neural network-based surrogate. 

An alternative strategy is to jointly train both the VAE and the surrogate. In this case, the property values of the training molecules need to be computed. An advantage of this approach is that the latent space can be informed of the desired properties explicitly, however, the generative model optimization becomes harder as there are more training objectives to balance (empirically we find the losses to be higher) and a lot more oracle compute is required. In this study we chose to use a surrogate that directly predicts using the latent space to simplify the optimization.

\section{Experiments}
\subsection{Setup}
All methods are trained on the ZINC-250K dataset. The parameters are optimized using the Adam optimizer \cite{kingma2017adam}, using a cosine-annealing learning rate schedule for one cycle \cite{loshchilov2017sgdr} for a total of 100K iterations. To prevent KL from vanishing across models with minimal tuning, a cyclic beta schedule was used while training the VAE \cite{fu2019cyclical}.
We follow the LIMO paper and use the following properties to evaluate the quality of generated optimized molecules.
 \begin{itemize}
     \item \textbf{SA}: Synthetic-accessibility, computed with SAScorer \cite{Ertl2009}. A score between 1 (easy to make) and 10 (very difficult to make).
     \item \textbf{QED}:  Quantitative Estimate of Drug-likeness, estimated by RDKit \cite{rdkit}. A value between 0 (non-drug-like) and 1 (drug-like).
     \item \textbf{BA}: Binding Affinity to the human estrogen receptor (PDB 1ERR), computed with AutoDock-GPU \cite{SantosMartins2021}. Outputs a prediction of the $\Delta G$ in kcal/mol, with lower values indicating stronger binding. We used 0 when the value could not be computed.
 \end{itemize}

 \begin{table}[t!]
\centering
\caption{Property comparison of all generated molecules from different model variants.}
\label{tab:all}
\begin{tabular}{cc|ccccc}
\toprule
\multicolumn{2}{c}{Model Variants}  & Count with valid AutoDock BA & $BA \downarrow$ & $SA \downarrow$ & $QED \uparrow$ \\
\midrule
\multirow{4}{*}{LIMO} & Default-Retrained & 9813 & -4.67 & 4.46 & 0.57 \\
& GS-Paper & 8729 & -5.44 & 2.89 & 0.62 \\
& GS-USPTO & 9947 & -5.56 & 4.46 & 0.61 \\
& GS-Zinc & 9868 & -5.57 & 4.63 & 0.59 \\
\midrule
\multirow{2}{*}{AR} & Default & 9194 & -5.03 & 4.93 & 0.56 \\
& Joint-Z & 9229 & -5.57 & 3.56 & 0.76 \\
\midrule
\multirow{3}{*}{AR-JointZ} & GS-Paper & 9768 & -5.26 & \textbf{2.91} & \textbf{0.81} \\
& GS-USPTO & 9939 & -5.58 & 3.28 & 0.73 \\
& GS-Zinc & 9967 & \textbf{-5.74} & 3.39 & 0.80 \\
\midrule
\multicolumn{2}{c|}{CMLMC} 
  & 8758 & -5.11 & 3.65 & 0.30 \\
\midrule
\midrule
\multicolumn{2}{c|}{ZINC250K (Training dataset )}   & 242591 & -5.35 & 3.05 & 0.73 \\
\bottomrule
\end{tabular}
\end{table}

\subsection{Property Optimization Results}
We apply reverse optimization to generate a set of random molecules from the trained generative model. For each experiment, the property optimization step was applied with 10,000 starting points and all of the resulting minima are considered. 

We consider the following model variants (1) LIMO: The original LIMO \cite{eckmann2022limo}.
(2) AR: LIMO with decoder replaced with an autoregressive Transformer model.
(3) AR-JointZ: AR with surrogate trained simultaneously with VAE.
(4) CMLMC: Non-Autoregressive decoder model from \cite{huang2022improving}
For LIMO, AR and AR-JointZ, we further consider different molecular string representation schemes (see Section \ref{molrep} for more details on the various molecular representations).

The results in Tables \ref{tab:all} show the number of binding ligands and the average property scores of those ligands for the human estrogen receptor (PDB 1ERR) target by considering 10,000 generated molecules each. We also report in Appendix Table \ref{tab:ba}, \ref{tab:sa}, \ref{tab:qed} the results for molecules with the top 100 BA, the molecules with the top 100 SA, and the molecules with the top 100 QED respectively.
The AR VAE decoder with the GroupSELFIES variants has the molecules with the best generated SA, QED (tokens provided by \cite{groupselfies2022}), and BA (tokens extracted from ZINC250K). 

We also include the corresponding property values of the training dataset (ZINC250K) as a reference. Note that the property values of generated molecules are even better than those of the training dataset. This demonstrates the great potential of using LIMO for de novo drug discovery.

\begin{figure}
    \centering
    \includegraphics[width=0.32\linewidth]{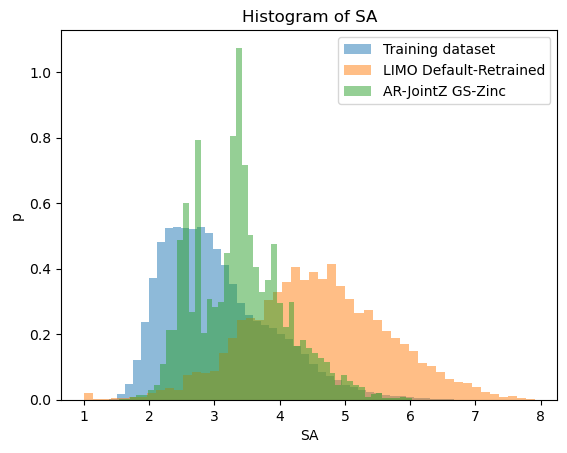}
\includegraphics[width=0.32\linewidth]{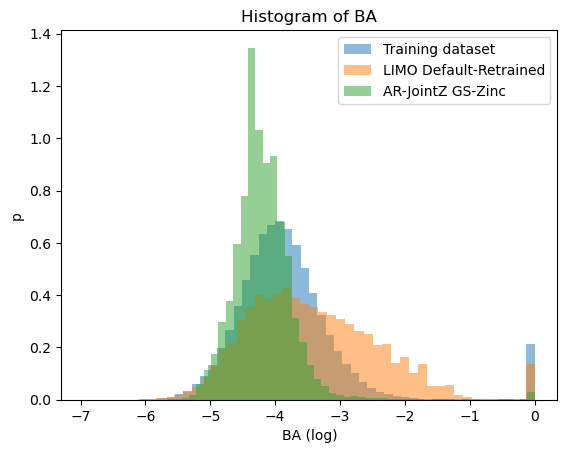}
\includegraphics[width=0.32\linewidth]{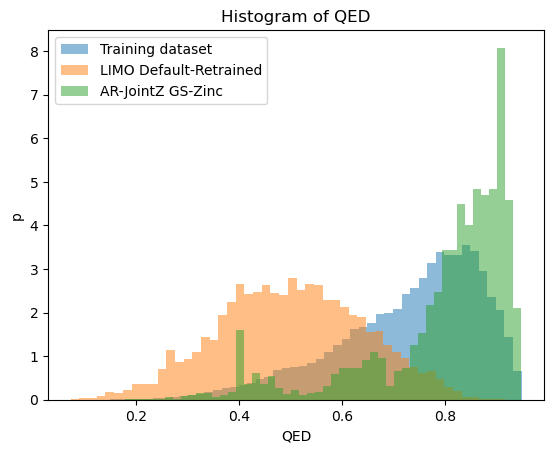}
    \caption{Distributions of property values for optimized molecules generated by LIMO, AR-JointZ GS-Zinc, and the training dataset.}
    \label{fig:dist}
\end{figure}

Figure~\ref{fig:dist} compares the distributions of property values for optimized molecules generated by LIMO, AR-JointZ GS-Zinc, and the training dataset. We can see that AR-JointZ with GS-ZINC representation improves the original LIMO model across all metrics. The model even outperforms the training dataset in terms of BA and QED values.

\subsection{Latent space analysis}
While Table \ref{tab:all} shows that the trained models are able to match the property distribution in the training dataset approximately, Table \ref{tab:ba}, \ref{tab:sa}, \ref{tab:qed} in Appendix show that the optimized molecules, however, fall short of molecules with the best property values in the training dataset. This difference in the best property values is due to the surrogate model-based optimization in the latent space. 

\begin{wrapfigure}{r}{0.5\linewidth} 
  \begin{center}
\includegraphics[width= \linewidth]{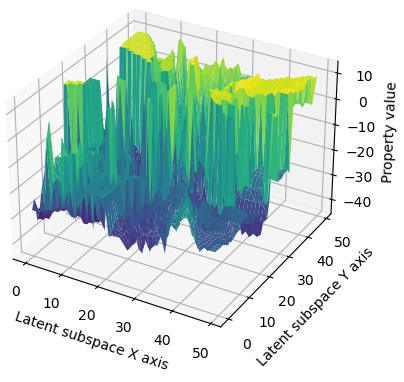}
    \caption{Objective plotted along a random subspace of the latent space. Latent space points were decoded to molecules by uniformly sampling a region constructed using two orthogonal random directions. The plot shows that objective values vary greatly even for nearby latent points.}
    \label{fig:loss_landscape}
      \end{center}
      \vspace{-10mm}
\end{wrapfigure}

We first investigate the organization of the latent space and identify features that would indicate the ease of optimization in this space. Figure~\ref{fig:loss_landscape}  shows the property value distribution in the learned latent space of LIMO. In the case of sequentially training the VAE and then the surrogate, the latent space receives no explicit feedback from the property values. Hence, the property values are not smoothly distributed, i.e. molecules decoded from the same neighborhood of the latent space can have large variations in their property values.

Since the surrogate is a simple approximation of the true latent space to property mapping, we measure the mean-squared error and the Dirichlet energy (Equation~\ref{eq:1}) to quantify the difference in surface smoothness, as well as the local correlation as measured using the pearson correlation coefficient to quantify the correctness of gradient directions in both the surfaces.

\begin{equation}\label{eq:1}
 \lambda_y=\frac{1}{N} y^{\mathrm{T}} L y
\end{equation}

where $y$ is the property value, $L=D-A$, $D$ is the degree matrix, $A$ is the KNN adjacency matrix, and $N$ is the number of samples considered for computing the Dirichlet energy. The property value measured in our experiments is the weighted sum of SA, QED, and BA as used in \cite{eckmann2022limo}. The training molecules are encoded into the latent space and 5 nearest neighbors are found using KNN which are then used to compute the Dirichlet energy.

The mean-squared error was found to decrease when training the surrogate jointly with the VAE (0.634) vs when training them separately (0.986), indicating that joint training leads to a better fit.

Additionally, the Dirichlet energies are lower when jointly training (33.31 vs 94.04), indicating that optimizing for properties using the jointly trained surrogate leads to smoother changes in property values i.e. molecules decoded from the same neighborhood of the latent space tend to have more similar property values.

Lastly, the local correlation as measured using the pearson correlation coefficient decreases when jointly training (0.48 vs 0.76), indicating that local gradient-based optimization using the jointly trained surrogate is less accurate.

\section{Future work}

\subsection{Latent space organization}
While jointly training the VAE and the surrogate improves the generated molecules by explicit feedback making the latent space interpolable with respect to properties, further improvements to the latent space organization via topological constraints \cite{moor2021topological, keller2022topographic} is a promising direction.
\subsection{Semi-autoregressive modelling}
There is a tradeoff between controllable generation and sequence modeling ability by changing the non-autoregressive decoder to an autoregressive one. \cite{huang2022improving} is an attempt at merging the best of both worlds by having an iterative non-autoregressive decoder. Other such approaches include semi-autoregressive models such as those used in \cite{han2023ssdlm} which generate blocks of text autoregressively, and order-agnostic autoregressive models such as those used in \cite{Alamdari2023} which can generate text in an arbitrary order i.e. not just left to right.  
\subsection{Further analysis of GroupSELFIES tokens}
The experiments done in this work show that the performance of the GroupSELFIES tokens depends greatly on the extracted tokens. While there is no obvious difference between the tokens extracted from different datasets, further study is needed to determine the cause of the apparent different in performance.

Pursuing these directions should improve the properties of molecules generated by LIMO, as well as maintain the ability to constrain the molecular scaffold.

\bibliography{iclr2021_conference}
\bibliographystyle{iclr2021_conference}
\newpage
\appendix
\section{Appendix}

We report the top 100 molecules in terms of SA, BA and QED in the corresponding tables.

\begin{table}[h!]
\centering
\caption{Top 100 BA molecules}
\label{tab:ba}
\begin{tabular}{cc|ccccc}
\midrule
\multicolumn{2}{c}{Model Variants}  & Count with valid AutoDock BA & $BA \downarrow$ & $SA \downarrow$ & $QED \uparrow$ \\
\midrule
\multirow{4}{*}{LIMO} & Default-Retrained & 9813 & -7.05 & 4.94 & 0.58 \\
& GS-Paper & 8729 & -7.65 & 3.54 & 0.49 \\
& GS-USPTO & 9947 & -7.55 & 4.75 & 0.59 \\
& GS-Zinc & 9868 & -7.46 & 4.97 & 0.57 \\
\midrule
\multirow{2}{*}{AR} & Default & 9194 & -6.80 & 5.18 & 0.67 \\
& Joint-Z & 9229 & -7.16 & 4.32 & 0.78 \\
\midrule
\multirow{3}{*}{AR-JointZ} & GS-Paper & 9768 & -6.53 & \textbf{3.05} & 0.82 \\
& GS-USPTO & 9939 & -7.11 & 3.41 & 0.77 \\
& GS-Zinc & 9967 & -7.11 & 3.65 & \textbf{0.84} \\
\midrule
CMLMC &  & 8758 & \textbf{-7.85} & 4.04 & 0.28 \\
\midrule
\midrule
Dataset & ZINC250K & 242591 & -8.22 & 3.08 & 0.51 \\
\bottomrule
\end{tabular}%
\end{table}

\begin{table}[h!]
\centering
\caption{Top 100 SA molecules}
\label{tab:sa}
\begin{tabular}{cc|ccccc}
\midrule
\multicolumn{2}{c}{Model Variants}  & Count with valid AutoDock BA & $BA \downarrow$ & $SA \downarrow$ & $QED \uparrow$ \\
\midrule
\multirow{4}{*}{LIMO} & Default-Retrained & 9813 & -3.24 & 2.19 & 0.52 \\
& GS-Paper & 8729 & -3.36 & \textbf{1.12} & 0.53 \\
& GS-USPTO & 9947 & -3.52 & 2.27 & 0.56 \\
& GS-Zinc & 9868 & -3.87 & 2.57 & 0.54 \\
\midrule
\multirow{2}{*}{AR} & Default & 9194 & -4.38 & 2.14 & 0.68 \\
& Joint-Z & 9229 & -5.02 & 1.80 & \textbf{0.85} \\
\midrule
\multirow{3}{*}{AR-JointZ} & GS-Paper & 9768 & -5.10 & 1.72 & 0.74 \\
& GS-USPTO & 9939 & -5.19 & 1.83 & \textbf{0.85} \\
& GS-Zinc & 9967 & \textbf{-5.37} & 2.01 & \textbf{0.85} \\
\midrule
CMLMC &  & 8758 & -4.40 & 1.74 & 0.22 \\
\midrule
\midrule
Dataset & ZINC250K & 242591 & -4.74 & 1.42 & 0.80 \\
\bottomrule
\end{tabular}%
\end{table}

\begin{table}[h]
\centering
\caption{Top 100 QED molecules}
\label{tab:qed}
\begin{tabular}{ccccccc}
\midrule
\multicolumn{2}{c}{Model Variants}  & Count with valid AutoDock BA & $BA \downarrow$ & $SA \downarrow$ & $QED \uparrow$ \\
\midrule
\multirow{4}{*}{LIMO} & Default-Retrained & 9813 & -5.49 & 4.65 & 0.83 \\
 & GS-Paper & 8729 & -5.47 & 3.02 & 0.91 \\
 & GS-USPTO & 9947 & -5.83 & 4.51 & 0.89 \\
 & GS-Zinc & 9868 & \textbf{-5.93} & 4.60 & 0.86 \\
\midrule
AR & Default & 9194 & -5.52 & 3.83 & 0.91 \\
 & Joint-Z & 9229 & -5.46 & 2.74 & \textbf{0.94} \\
\midrule
\multirow{3}{*}{AR-JointZ} & GS-Paper & 9768 & -5.23 & \textbf{2.55} & \textbf{0.94} \\
 & GS-USPTO & 9939 & -5.54 & 2.57 & \textbf{0.94} \\
 & GS-Zinc & 9967 & -5.67 & 2.60 & \textbf{0.94} \\
\midrule
CMLMC &  & 8758 & -5.43 & 4.59 & 0.77 \\
\midrule
\midrule
Dataset & ZINC250K & 242591 & -5.38 & 2.75 & 0.95 \\
\bottomrule
\end{tabular}%
\end{table}


\end{document}

%% file: math_commands.tex
\usepackage{amsmath,amsfonts,bm}

\def\eqref#1{equation~\ref{#1}}

\def\1{\bm{1}}

\DeclareMathAlphabet{\mathsfit}{\encodingdefault}{\sfdefault}{m}{sl}
\SetMathAlphabet{\mathsfit}{bold}{\encodingdefault}{\sfdefault}{bx}{n}

